\documentclass[11pt]{article}

\usepackage{acl}
\usepackage{times}
\usepackage{latexsym}
\usepackage[T1]{fontenc}
\usepackage[utf8]{inputenc}
\usepackage{microtype}
\usepackage{inconsolata}
\usepackage{graphicx}
\usepackage{amsmath}
\usepackage{amssymb}
\usepackage{booktabs}
\usepackage{multirow}
\usepackage{subcaption}
\usepackage{array}
\title{Benchmark$^2$: Systematic Evaluation of LLM Benchmarks}

\begin{document}

\author{
  \textbf{Qi Qian}$^{1*}$ \quad
  \textbf{Chengsong Huang}$^{2*}$ \quad
  \textbf{Jingwen Xu}$^{1}$ \quad
  \textbf{Changze Lv}$^{1}$ \quad
  \textbf{Muling Wu}$^{1}$ \quad
  \textbf{Wenhao Liu}$^{3}$ \\
  \textbf{Xiaohua Wang}$^{1}$ \quad
  \textbf{Zhenghua Wang}$^{1}$ \quad
  \textbf{Zisu Huang}$^{1}$ \quad
  \textbf{Muzhao Tian}$^{1}$ \quad
  \textbf{Jianhan Xu}$^{3}$ \quad
  \textbf{Kun Hu}$^{3}$ \\
  \textbf{He-Da Wang}$^{3}$ \quad
  \textbf{Yao Hu}$^{3}$ \quad
  \textbf{Xuanjing Huang}$^{1\dagger}$ \quad
  \textbf{Xiaoqing Zheng}$^{1\dagger}$ \\[0.5em]
  $^1$College of Computer Science and Artificial Intelligence, Fudan University \\
  $^2$Washington University in St. Louis \quad
  $^3$Xiaohongshu Inc. \\[0.3em]
  \texttt{qqian23@m.fudan.edu.cn\quad chengsong@wustl.edu} \\
  \texttt{\{zhengxq, xjhuang\}@fudan.edu.cn} \\[0.3em]
}

\maketitle

\begin{abstract}
The rapid proliferation of benchmarks for evaluating large language models (LLMs) has created an urgent need for systematic methods to assess benchmark quality itself. We propose \textsc{Benchmark}$^2$, a comprehensive framework comprising three complementary metrics: (1) \textbf{Cross-Benchmark Ranking Consistency}, measuring whether a benchmark produces model rankings aligned with peer benchmarks; (2) \textbf{Discriminability Score}, quantifying a benchmark's ability to differentiate between models; and (3) \textbf{Capability Alignment Deviation}, identifying problematic instances where stronger models fail but weaker models succeed \textit{within the same model family}. We conduct extensive experiments across 15 benchmarks spanning mathematics, reasoning, and knowledge domains, evaluating 11 LLMs across four model families. Our analysis reveals significant quality variations among existing benchmarks and demonstrates that selective benchmark construction based on our metrics can achieve comparable evaluation performance with substantially reduced test sets.
\end{abstract}

\section{Introduction}

The evaluation of large language models (LLMs) has become increasingly important as these models are deployed across diverse real-world applications. Benchmarks serve as the primary instruments for measuring model capabilities, guiding both research directions and practical deployment decisions. However, the explosive growth in the number of benchmarks—with hundreds now available across domains such as mathematics, reasoning, instruction following, and knowledge understanding—raises a fundamental question: \textit{How do we know if a benchmark itself is good?}

\begin{figure*}[t]
    \centering
    \includegraphics[width=\textwidth]{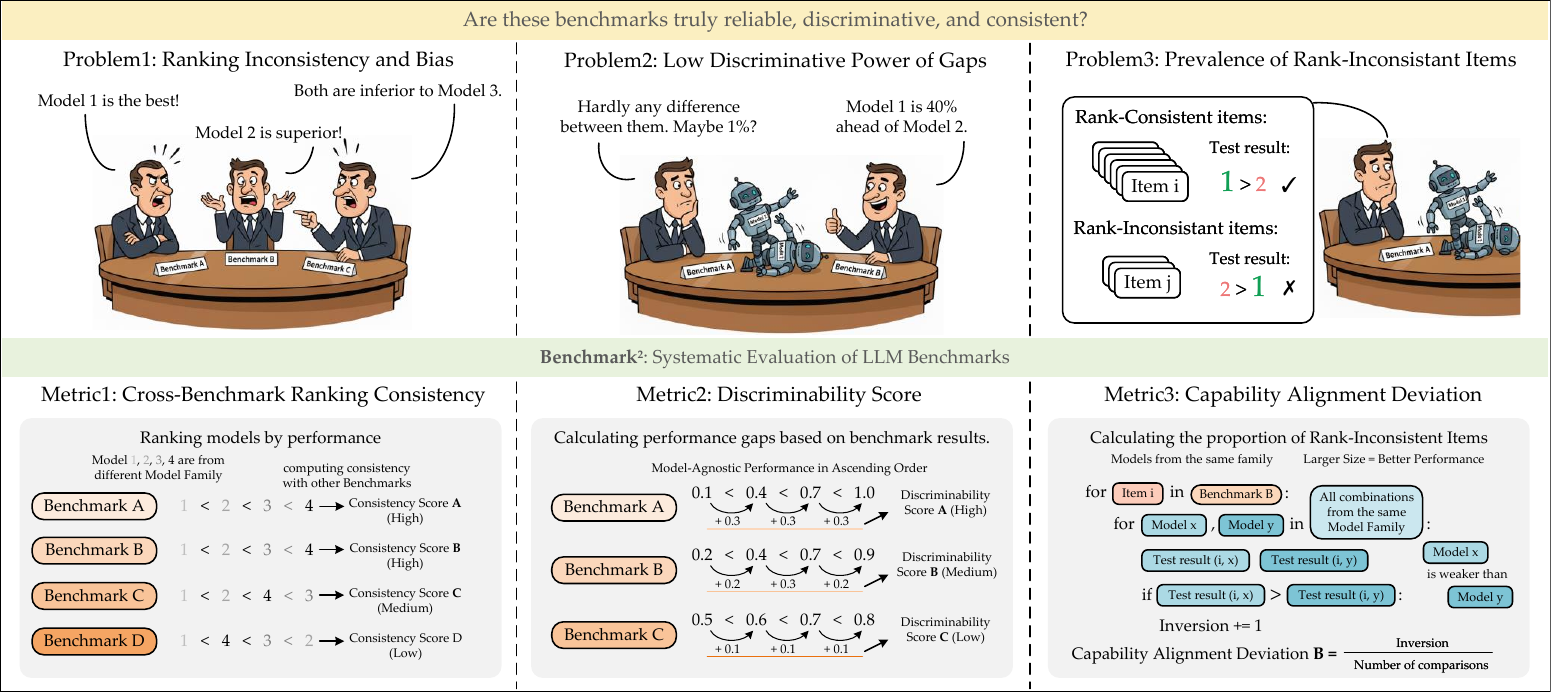}
    \caption{Overview of \textsc{Benchmark}$^2$ framework. \textbf{Top row}: Three key problems with existing LLM benchmarks—ranking inconsistency across benchmarks, low discriminative power of performance gaps, and prevalence of rank-inconsistent test items. \textbf{Bottom row}: Our three complementary metrics addressing each problem—Cross-Benchmark Ranking Consistency (CBRC) measures alignment with peer benchmarks, Discriminability Score (DS) quantifies performance gap magnitudes, and Capability Alignment Deviation (CAD) identifies items violating expected capability hierarchies within model families.}
    \label{fig:framework_overview}
\end{figure*}

Although benchmarks serve an important role in the field, surprisingly little attention has been paid to the benchmark quality. Current practice often treats benchmarks as ground truth without questioning their reliability or validity. This oversight can lead to several problems, as illustrated in Figure~\ref{fig:framework_overview}: (1) \textbf{Ranking Inconsistency}—different benchmarks may produce conflicting model rankings, making it unclear which benchmark to trust; (2) \textbf{Low Discriminative Power}—some benchmarks fail to differentiate between models of varying capabilities, clustering all models within a narrow performance range; and (3) \textbf{Rank-Inconsistent Items}—individual test instances may exhibit counter-intuitive behavior where stronger models fail but weaker models succeed.

Consider a concrete example: if Benchmark A ranks Model X above Model Y in mathematical reasoning, while Benchmarks B and C consistently show the opposite ranking, should we trust Benchmark A? Similarly, if a benchmark shows minimal performance differences between a state-of-the-art model and a much smaller model, does this indicate that these two models have similar capabilities, or does it suggest that the benchmark cannot effectively distinguish between them?

To address these challenges, we propose \textsc{Benchmark}$^2$, a novel framework for evaluating benchmark quality through three complementary metrics (Figure~\ref{fig:framework_overview}, bottom row):
\textbf{Cross-Benchmark Ranking Consistency (CBRC)}, which measures ranking correlation with external domain benchmarks, \textbf{Discriminability Score (DS)}, which measures the benchmark's ability to distinguish between models of varying capabilities; and \textbf{Capability Alignment Deviation (CAD)}, which penalizes counter-intuitive instances where weaker models outperform stronger ones within the same family, ensuring hierarchical consistency.



We conduct comprehensive experiments across 15 widely-used benchmarks spanning three major domains (mathematics, general reasoning, and knowledge \& understanding), evaluating 11 LLMs across four model families with clear capability hierarchies. Our analysis reveals substantial quality variations among existing benchmarks and identifies specific characteristics that distinguish high-quality benchmarks from problematic ones.

Beyond benchmark quality assessment, we demonstrate a practical application of our framework: selective benchmark construction. By identifying high-quality test instances based on our metrics, we construct reduced benchmark versions that achieve comparable evaluation performance to full benchmarks while providing greater efficiency.

Our contributions are summarized as follows:

\begin{itemize}
\setlength{\itemsep}{0pt}
    \setlength{\parsep}{0pt}
    \setlength{\parskip}{0pt}
    \setlength{\topsep}{0pt}
    \setlength{\partopsep}{0pt}
    \item We formalize the problem of benchmark quality assessment and propose three complementary metrics that capture different aspects of benchmark reliability.
    \item We conduct the first large-scale systematic evaluation of benchmark quality across 15 benchmarks and 11 models spanning four families, providing empirical insights into the state of LLM evaluation.
    \item We show that filtering instances via quality metrics achieves comparable evaluation performance using only 35\% of the original data.
\end{itemize}

\section{Related Work}

\subsection{LLM Benchmarks}

The landscape of LLM evaluation has expanded dramatically. General-purpose benchmarks such as MMLU \cite{hendrycks2021mmlu}, BBH \cite{suzgun2023bbh}, and ARC \cite{clark2018arc} measure broad capabilities across multiple domains. Domain-specific benchmarks have emerged for mathematics (MATH-500, AIME, OlympiadBench), reasoning and comprehension (DROP~\citep{dua2019drop}, CommonsenseQA~\citep{talmor2019commonsenseqa}), and knowledge understanding (IFEval\cite{zhou2023ifeval}, SuperGPQA~\citep{mapteam2025supergpqa}). More challenging benchmarks have been introduced to address ceiling effects, including OmniMath~\citep{gao2024omnimath} for advanced mathematics. However, this proliferation has occurred largely without systematic quality assessment.

\subsection{Benchmark Quality Analysis}

Several studies have examined issues with existing benchmarks. \citet{bowman2021dangers} discussed dangers of benchmark-driven research and the need for more robust evaluation practices. Data contamination, where LLMs inadvertently encounter test data during training, has been identified as a significant concern \cite{xu2024contamination,sainz2023contamination}. Benchmark saturation has motivated dynamic benchmarks \cite{kiela2021dynabench}. Research on evaluation methodology has addressed statistical significance in model comparisons \cite{dror2018hitchhiker} and limitations of single-number metrics \cite{ethayarajh2020utility}. \citet{liang2023helm} introduced HELM for holistic evaluation across multiple dimensions.

Our work complements these efforts by providing metrics specifically designed for benchmark quality assessment. Unlike prior work that focuses on identifying specific issues or proposing new evaluation paradigms, our framework provides systematic, quantitative metrics for assessing benchmark reliability, discriminability, and capability alignment.

\section{Methodology}
In this section, we present \textsc{Benchmark}$^2$, our framework for benchmark evaluation.
Formally, consider a set of benchmarks $\mathcal{B} = \{B_1, B_2, \ldots, B_n\}$ and a set of candidate models $\mathcal{M} = \{M_1, M_2, \ldots, M_m\}$. 
Let $s_{ij}$ denote the performance score of model $M_j$ evaluated on benchmark $B_i$. 
Based on this formulation, we propose three complementary approaches to assess benchmark quality:

\subsection{Cross-Benchmark Ranking Consistency}

Cross-Benchmark Ranking Consistency (CBRC) evaluates whether a benchmark's ranking corroborates with others in the same domain. The underlying rationale is that effective benchmarks measuring similar capabilities should produce highly correlated model rankings.

\paragraph{Definition.} For a benchmark $B_i$, we compute its ranking consistency as the average Kendall's $\tau$ correlation with other benchmarks in the same domain:

\begin{equation}
    \text{CBRC}(B_i) = \frac{1}{n-1} \sum_{j \neq i} \tau(r_i, r_j)
\end{equation}

where $r_i$ denotes the ranking of models induced by benchmark $B_i$, and $\tau(\cdot, \cdot)$ is Kendall's tau correlation coefficient.

\paragraph{Interpretation.} CBRC values range from -1 to 1, where 1 indicates perfect agreement with other benchmarks, 0 indicates no correlation, and negative values indicate inverse rankings. We consider CBRC $> 0.7$ as indicating high consistency, $0.4$--$0.7$ as moderate, and $< 0.4$ as low.

\subsection{Discriminability Score}

A high-quality benchmark should effectively differentiate between models of varying capabilities. If all models achieve similar scores regardless of their actual ability differences, the benchmark provides limited useful information.

\paragraph{Definition.} We define the Discriminability Score (DS) based on the normalized score spread and the statistical significance of pairwise differences:

\begin{equation}
    \text{DS}(B_i) = \frac{\sigma_i}{\bar{s}_i} \cdot \sqrt{\frac{\sum_{j < k} \mathbf{1}[|s_{ij} - s_{ik}| > \epsilon]}{m(m-1)/2}}
\end{equation}

where $\sigma_i$ is the standard deviation of scores on benchmark $B_i$, $\bar{s}_i$ is the mean score, and the second term represents the proportion of model pairs with practically significant differences (we set $\epsilon = 0.02$ as the minimum meaningful difference).

\paragraph{Interpretation.} Higher DS values indicate better discriminability. We empirically find that benchmarks with DS $> 0.4$ provide good differentiation, while those with DS $< 0.2$ offer minimal discrimination between models.

\subsection{Capability Alignment Deviation}

This metric operates at the instance level, identifying individual test questions that may be problematic. The key insight is that if a question is well-designed, stronger models should generally outperform weaker models on it, maintaining alignment with expected capability hierarchies.

\paragraph{Model Family Hierarchy.} Rather than establishing a global ordering across all models—which can be unreliable due to different training data and optimization objectives across model families—we leverage the natural capability hierarchy \textit{within} model families. For a model family $\mathcal{F}$ (e.g., Qwen2.5-Instruct), we define an ordering based on parameter count:
\begin{equation}
    \mathcal{F} = \{M_1 \succ M_2 \succ \cdots \succ M_k\}
\end{equation}
where $M_1$ has the most parameters and $M_k$ has the fewest. This within-family ordering is more reliable than cross-family comparisons.

\paragraph{Definition.} For a benchmark $B_i$, we first compute the raw inversion rate by aggregating inversions across all model families $\{\mathcal{F}_1, \mathcal{F}_2, \ldots, \mathcal{F}_F\}$:
\begin{equation}
    \text{inv\_rate}(B_i) = \frac{\sum_{f=1}^{F} \text{inv}_{\mathcal{F}_f}(B_i)}{\sum_{f=1}^{F} \text{comp}_{\mathcal{F}_f}(B_i)}
\end{equation}

For each family $\mathcal{F} = \{M_1 \succ M_2 \succ \cdots \succ M_k\}$, an inversion on question $q$ occurs when a stronger model fails but a weaker model succeeds:
\begin{equation}
    \text{inv}_{\mathcal{F}}(B_i) = \sum_{q \in Q_i} \sum_{j < l} \mathbf{1}[\neg c_{jq} \land c_{lq}]
\end{equation}
where $Q_i$ is the set of questions in benchmark $B_i$, $c_{jq}$ indicates whether model $M_j$ correctly answers question $q$.

We then apply an exponential transformation to convert the inversion rate to a score where higher values indicate better alignment:
\begin{equation}
    \text{CAD}(B_i) = e^{-\lambda \cdot \text{inv\_rate}(B_i)}
\end{equation}
where $\lambda > 0$ is a scaling parameter that controls the sensitivity of the transformation. In our experiments, we set $\lambda = 12$ based on empirical analysis to ensure meaningful differentiation across the observed range of inversion rates.

\paragraph{Interpretation.} CAD ranges from 0 to 1, where 1 indicates perfect alignment (no inversions) and values approaching 0 indicate severe capability hierarchy violations. We consider CAD $> 0.6$ as indicating good quality, $0.4$--$0.6$ as acceptable, and $< 0.4$ as indicating significant quality issues.

\subsection{Stability Score}

To assess the reliability of selective benchmark evaluation, we introduce the Stability Score, which measures the consistency of model rankings across multiple sampling iterations.

\paragraph{Definition.} For a selective benchmark $B_s$ with selection ratio $r$, we perform $K$ bootstrap sampling iterations (we use $K=100$). In each iteration $k$, we sample $r \cdot |B|$ instances and compute the resulting model ranking $r_k$. The Stability Score is defined as the average pairwise ranking correlation:

\begin{equation}
    \text{Stability}(B_s) = \frac{2}{K(K-1)} \sum_{i < j} \tau(r_i, r_j)
\end{equation}

where $\tau(\cdot, \cdot)$ is Kendall's tau correlation coefficient between rankings from different bootstrap samples.

\paragraph{Interpretation.} Stability Score ranges from -1 to 1, where 1 indicates that the selective benchmark produces identical rankings regardless of which specific instances are sampled, and lower values indicate higher variance in rankings. We consider Stability $> 0.7$ as high, $0.5$--$0.7$ as moderate, and $< 0.5$ as low.

\subsection{Benchmark Quality Score}

We also provide a combined score for overall assessment:

\begin{equation}
\begin{split}
    \text{BQS}(B_i) = \alpha \cdot \widetilde{\text{CBRC}}(B_i) &+ \beta \cdot \text{DS}(B_i) \\
    &+ \gamma \cdot \text{CAD}(B_i)
\end{split}
\end{equation}
\vspace{-3pt}

where $\widetilde{\text{CBRC}}$ denotes the normalized CBRC score, and $\alpha$, $\beta$, $\gamma$ are weighting parameters. Details on normalization and weight selection are provided in Appendix~\ref{sec:appendix_bqs}.

\begin{table*}[t]
\centering
\small
\setlength{\tabcolsep}{8pt}
\begin{tabular}{l cc ccc cc w{c}{1.2cm}}
\toprule
\multirow{2}{*}{\textbf{Benchmark}} & \multicolumn{2}{c}{\textbf{CBRC}} & \multicolumn{3}{c}{\textbf{DS}} & \multicolumn{2}{c}{\textbf{CAD}} & \multirow{2}{*}{\textbf{BQS}} \\
\cmidrule(lr){2-3} \cmidrule(lr){4-6} \cmidrule(lr){7-8}
 & Score & $\sigma$ & Score & Range & $\sigma$ & Score & $\sigma$ & \\
\midrule
\multicolumn{9}{c}{\textit{Mathematics}} \\
\midrule
AIME 2024 & 0.52 & 0.10 & 0.74 & 0--53 & 0.22 & 0.85 & 0.13 & 0.79 \\
OmniMath & 0.76 & 0.13 & 0.79 & 0--62 & 0.27 & 0.61 & 0.22 & 0.75 \\
OlympiadBench & 0.75 & 0.12 & 0.76 & 0--64 & 0.32 & 0.61 & 0.23 & 0.73 \\
AMC 22-24 & 0.70 & 0.18 & 0.36 & 16--67 & 0.11 & 0.46 & 0.09 & 0.55 \\
MATH-500 & 0.70 & 0.18 & 0.16 & 49--87 & 0.07 & 0.62 & 0.10 & 0.55 \\
\midrule
\multicolumn{9}{c}{\textit{General Reasoning}} \\
\midrule
ARC & 0.79 & 0.03 & 0.11 & 56--95 & 0.07 & 0.87 & 0.05 & 0.65 \\
BBH & 0.75 & 0.02 & 0.25 & 30--89 & 0.12 & 0.66 & 0.05 & 0.60 \\
DROP & 0.71 & 0.08 & 0.20 & 41--87 & 0.05 & 0.61 & 0.07 & 0.56 \\
CommonsenseQA & 0.75 & 0.07 & 0.17 & 37--85 & 0.11 & 0.57 & 0.02 & 0.54 \\
SIQA & 0.73 & 0.05 & 0.17 & 24--53 & 0.09 & 0.23 & 0.03 & 0.40 \\
\midrule
\multicolumn{9}{c}{\textit{Knowledge \& Understanding}} \\
\midrule
IFEval & 0.75 & 0.11 & 0.23 & 37--87 & 0.10 & 0.63 & 0.10 & 0.58 \\
EQ-Bench & 0.75 & 0.08 & 0.27 & 17--82 & 0.17 & 0.53 & 0.07 & 0.56 \\
IFBench & 0.71 & 0.11 & 0.31 & 11--32 & 0.08 & 0.51 & 0.07 & 0.55 \\
SuperGPQA & 0.79 & 0.05 & 0.34 & 9--40 & 0.08 & 0.43 & 0.07 & 0.55 \\
MMLU-Pro & 0.65 & 0.10 & 0.40 & 27--71 & 0.17 & 0.36 & 0.18 & 0.51 \\
\bottomrule
\end{tabular}
\caption{Comprehensive benchmark quality metrics across three domains. \textbf{CBRC}: Cross-Benchmark Ranking Consistency (Kendall's $\tau$ with peer benchmarks). \textbf{DS}: Discriminability Score; Range shows model performance spread (min--max \%). \textbf{CAD}: Capability Alignment Deviation (higher is better). \textbf{BQS}: Combined Benchmark Quality Score. \textbf{$\sigma$} denotes standard deviation.}
\label{tab:comprehensive_quality}
\end{table*}

\section{Experimental Setup}

\subsection{Benchmarks}
To demonstrate the broad applicability of our framework, we select a diverse set of 15 benchmarks across three major domains:
\paragraph{Mathematics (5 benchmarks).} AIME 2024 \cite{aime2024}, OmniMath \cite{gao2024omnimath} for advanced mathematical problem solving, OlympiadBench \cite{he2024olympiadbench}, AMC \cite{amc2024} and MATH-500 \cite{hendrycks2021math}.
\paragraph{General Reasoning (5 benchmarks).} Big-Bench Hard (BBH)~\cite{suzgun2023bbh} for challenging tasks, DROP \cite{dua2019drop} for reading comprehension, ARC \cite{clark2018arc} for scientific reasoning, CommonsenseQA \cite{talmor2019commonsenseqa} for commonsense reasoning and SIQA \cite{sap2019siqa} for social intelligence.
\paragraph{Knowledge \& Understanding (5 benchmarks).} SuperGPQA \cite{mapteam2025supergpqa} for graduate-level reasoning, MMLU-Pro \cite{wang2024mmlupro} , IFBench~\cite{pyatkin2025ifbench} and IFEval \cite{zhou2023ifeval} for instruction following capabilities and EQ-Bench \cite{paech2023eqbench} for emotional intelligence.

\subsection{Models}

We evaluate 11 models across four families with clear capability hierarchies based on model size:
\begin{itemize}
\setlength{\itemsep}{0pt}
    \setlength{\itemsep}{0pt}
    \setlength{\parsep}{0pt}
    \setlength{\parskip}{0pt}
    \setlength{\topsep}{0pt}
    \setlength{\partopsep}{0pt}
    \item \textbf{DeepSeek-R1-Distill-Qwen~\citep{deepseekai2025deepseekr1incentivizingreasoningcapability}}: 1.5B, 7B, 32B
    \item \textbf{Llama-3.1-Instruct~\citep{grattafiori2024llama}}: 8B, 70B
    \item \textbf{Qwen2.5-Instruct~\citep{Yang2024Qwen25TR}}: 1.5B, 7B, 72B
    \item \textbf{Qwen3~\citep{yang2025qwen3}}: 1.7B, 8B, 32B
\end{itemize}

\paragraph{Selection Rationale.} To ensure architectural diversity and enable reliable CAD computation across multiple scales, we selected models from four distinct development lineages. This strategy mitigates the risk of family-specific bias while providing 1--3 comparison pairs within each family, yielding a total of 10 pairs per benchmark instance.

\paragraph{Held-Out Validation.} To verify that our metrics generalize beyond the models used in their computation, we additionally evaluate on \textbf{Qwen2.5-Base} (1.5B, 7B, 32B)—base models that share the same architecture as Qwen2.5-Instruct but were \textbf{not used} in metric computation and have fundamentally different training (no instruction tuning).

\subsection{Evaluation Protocol}

For each model-benchmark pair, we use standardized prompting templates following the original benchmark specifications where available. We use greedy decoding for reproducibility and evaluate using exact match or execution-based metrics as appropriate for each benchmark.

\begin{table*}[t]
\centering
\setlength{\tabcolsep}{3pt}
\small
\begin{tabular}{l cccc cccc cccc cccc}
\toprule
\multirow{2}{*}{\textbf{Model}} & \multicolumn{4}{c}{\textbf{Mathematics}} & \multicolumn{4}{c}{\textbf{General}} & \multicolumn{4}{c}{\textbf{K\&U}} & \multicolumn{4}{c}{\textbf{Avg}} \\
\cmidrule(lr){2-5} \cmidrule(lr){6-9} \cmidrule(lr){10-13} \cmidrule(lr){14-17} 
& F & S & $\Delta$ & Rk & F & S & $\Delta$ & Rk & F & S & $\Delta$ & Rk & F & S & $\Delta$ & Rk\\
\midrule
\multicolumn{17}{c}{\textit{DeepSeek-R1-Distill-Qwen}} \\
\midrule
32B & 57.5 & 80.1 & +22.6 & 3$\to$3 & 78.8 & 85.5 & +6.7 & 3$\to$3 & 52.8 & 67.5 & +14.7 & 5$\to$4 & 63.1 & 77.7 & +14.6 & 4$\to$3 \\
7B & 42.8 & 57.7 & +14.9 & 5$\to$5 & 63.9 & 47.7 & -16.2 & 10$\to$10 & 36.6 & 22.1 & -14.5 & 10$\to$11 & 47.8 & 42.5 & -5.3 & 8$\to$9 \\
1.5B & 26.9 & 26.0 & -0.9 & 10$\to$10 & 38.3 & 13.5 & -24.8 & 13$\to$13 & 20.4 & 6.1 & -14.3 & 13$\to$13 & 28.5 & 15.2 & -13.3 & 13$\to$13 \\
\midrule
\multicolumn{17}{c}{\textit{Llama-3.1-Instruct}} \\
\midrule
70B & 30.1 & 32.1 & +2.0 & 8$\to$9 & 79.8 & 86.5 & +6.7 & 2$\to$2 & 60.9 & 79.8 & +18.9 & 3$\to$3 & 56.9 & 66.1 & +9.2 & 5$\to$5 \\
8B & 13.3 & 4.2 & -9.1 & 13$\to$13 & 67.9 & 56.6 & -11.3 & 8$\to$9 & 47.5 & 54.2 & +6.7 & 7$\to$8 & 42.9 & 38.3 & -4.6 & 10$\to$10 \\
\midrule
\multicolumn{17}{c}{\textit{Qwen2.5-Instruct}} \\
\midrule
72B & 49.4 & 79.3 & +29.9 & 4$\to$4 & 78.7 & 85.0 & +6.3 & 4$\to$4 & 62.0 & 81.3 & +19.3 & 1$\to$2 & 63.4 & 81.9 & +18.5 & 3$\to$2 \\
7B & 29.5 & 37.7 & +8.2 & 9$\to$8 & 73.3 & 69.4 & -3.9 & 6$\to$6 & 51.4 & 58.3 & +6.9 & 6$\to$6 & 51.4 & 55.1 & +3.7 & 6$\to$6 \\
1.5B & 13.7 & 6.2 & -7.5 & 12$\to$12 & 54.3 & 30.0 & -24.3 & 11$\to$12 & 29.8 & 20.1 & -9.7 & 11$\to$12 & 32.6 & 18.8 & -13.8 & 11$\to$12 \\
\midrule
\multicolumn{17}{c}{\textit{Qwen3}} \\
\midrule
32B & 63.5 & 97.1 & +33.6 & 1$\to$1 & 81.7 & 94.2 & +12.5 & 1$\to$1 & 61.9 & 85.7 & +23.8 & 2$\to$1 & 69.1 & 92.3 & +23.2 & 1$\to$1 \\
8B & 58.4 & 82.9 & +24.5 & 2$\to$2 & 78.3 & 82.9 & +4.6 & 5$\to$5 & 56.3 & 60.6 & +4.3 & 4$\to$5 & 64.3 & 75.5 & +11.2 & 2$\to$4 \\
1.7B & 34.9 & 49.7 & +14.8 & 6$\to$6 & 69.6 & 57.5 & -12.1 & 7$\to$8 & 44.8 & 37.2 & -7.6 & 8$\to$9 & 49.8 & 48.1 & -1.7 & 7$\to$8 \\
\midrule
\textbf{$\tau$} & \multicolumn{4}{c}{0.96} & \multicolumn{4}{c}{0.96} & \multicolumn{4}{c}{0.85} & \multicolumn{4}{c}{0.93} \\
\textbf{Stability} & \multicolumn{4}{c}{0.76} & \multicolumn{4}{c}{0.73} & \multicolumn{4}{c}{0.58} & \multicolumn{4}{c}{0.69} \\
\bottomrule
\end{tabular}
\caption{Model performance comparison between full benchmarks (F) and selective evaluation (S). $\Delta$ = score difference, Rk = rank change among all 14 models (Full$\to$Selective). Bottom rows show ranking consistency (Kendall's $\tau$) and stability score per domain.}
\label{tab:model_performance}
\end{table*}

\begin{table*}[t]
\centering
\setlength{\tabcolsep}{3pt}
\small
\begin{tabular}{l cccc cccc cccc cccc}
\toprule
\multirow{2}{*}{\textbf{Model}} & \multicolumn{4}{c}{\textbf{Mathematics}} & \multicolumn{4}{c}{\textbf{General}} & \multicolumn{4}{c}{\textbf{K\&U}} & \multicolumn{4}{c}{\textbf{Avg}} \\
\cmidrule(lr){2-5} \cmidrule(lr){6-9} \cmidrule(lr){10-13} \cmidrule(lr){14-17}
 & F & S & $\Delta$ & Rk & F & S & $\Delta$ & Rk & F & S & $\Delta$ & Rk & F & S & $\Delta$ & Rk \\
\midrule
32B & 34.5 & 45.0 & +10.5 & 7$\to$7 & 67.5 & 61.6 & -5.9 & 9$\to$7 & 39.7 & 56.7 & +17.0 & 9$\to$7 & 47.2 & 54.4 & +7.2 & 9$\to$7 \\
7B & 15.2 & 6.4 & -8.8 & 11$\to$11 & 52.8 & 42.8 & -10.0 & 12$\to$11 & 23.8 & 30.4 & +6.6 & 12$\to$10 & 30.6 & 26.5 & -4.1 & 12$\to$11 \\
1.5B & 1.5 & 1.1 & -0.4 & 14$\to$14 & 5.8 & 7.8 & +2.0 & 14$\to$14 & 4.2 & 5.3 & +1.1 & 14$\to$14 & 3.9 & 4.7 & +0.8 & 14$\to$14 \\
\midrule
\textbf{Avg $|\Delta\text{Rk}|$} & \multicolumn{4}{c}{0.0} & \multicolumn{4}{c}{1.0} & \multicolumn{4}{c}{1.3} & \multicolumn{4}{c}{1.0} \\
\bottomrule
\end{tabular}
\caption{Held-out model validation using \textbf{Qwen2.5-Base} family (1.5B, 7B, 32B). These base models were \textbf{not used} in computing CAD, DS, or CBRC metrics, which were derived exclusively from instruction-tuned models. F = Full benchmark score (\%), S = Selective evaluation score (\%), $\Delta$ = score difference, Rk = rank among all 14 models (Full$\to$Selective). Avg $|\Delta\text{Rk}|$ shows the average absolute rank change for held-out models (lower is better).}
\label{tab:held_out_validation}
\end{table*}

\section{Results}

\subsection{Comprehensive Quality Analysis}

Table~\ref{tab:comprehensive_quality} reveals distinct quality profiles across domains.

\textbf{Mathematics} exhibits the widest quality variation (BQS: 0.55--0.79). AIME 2024 achieves exceptional discriminability (DS = 0.74) and capability alignment (CAD = 0.85), while MATH-500 shows potential ceiling effects with low discriminability (DS = 0.16).

\textbf{General Reasoning} presents a quality-discriminability trade-off. ARC achieves the highest capability alignment (CAD = 0.87) but limited discriminability (DS = 0.11), whereas BBH maximizes discriminability (DS = 0.25) at the cost of alignment (CAD = 0.66). SIQA exhibits problematic quality across all model families (CAD = 0.23).

\textbf{Knowledge \& Understanding} shows the most consistent quality profile (BQS: 0.51--0.58), with IFEval and SuperGPQA achieving strong cross-benchmark consistency (CBRC $\geq$ 0.75).

Two patterns emerge: (1) high discriminability and high capability alignment rarely co-occur, and (2) benchmarks with objective evaluation criteria consistently achieve higher CAD scores.

\subsection{Model Performance Analysis}
Table~\ref{tab:model_performance} reveals clear capability hierarchies within each model family, validating our within-family CAD computation approach. Within the DeepSeek family, performance scales consistently from 28.5\% (1.5B) to 47.8\% (7B) to 63.1\% (32B) average. Similar patterns hold for Llama (42.9\% to 56.9\%), Qwen2.5 (32.6\% to 51.4\% to 63.4\%), and Qwen3 (49.8\% to 64.3\% to 69.1\%).

The selective evaluation results demonstrate that our quality-based instance selection maintains strong ranking consistency with full benchmarks. The average Kendall's $\tau$ of 0.93 indicates that selective evaluation preserves the relative ordering of models while using only 35\% of the original instances. Notably, larger models within each family consistently show positive $\Delta$ values on selective benchmarks, confirming that high-quality instances better differentiate capable models. The rank change column (Rk) shows that most models maintain similar rankings between full and selective evaluation, with only minor position shifts occurring primarily among mid-tier models.

\subsection{Held-Out Model Validation}
To validate that our metrics generalize beyond the models used in their computation, we evaluate on held-out models that were not included in computing CAD, DS, or CBRC metrics. Specifically, we use Qwen2.5-Base (1.5B, 7B, 32B), which are base models without instruction tuning. Table~\ref{tab:held_out_validation} presents these results.

The held-out validation demonstrates strong generalization of our selective benchmark approach. Mathematics shows perfect rank preservation (Avg $|\Delta\text{Rk}|$ = 0.0), indicating that selective evaluation ranks held-out models identically to full benchmarks in this domain. General Reasoning and Average scores show moderate variation (1.0), while Knowledge \& Understanding shows slightly higher variation (1.3). Notably, the extreme-performing 1.5B model maintains its 14th rank consistently across all domains, demonstrating that our selection method is particularly reliable at the capability distribution tails.

\subsection{Selective Benchmark Construction}

We select test instances with high CAD scores (indicating low inversion rates) and high discriminability contributions, creating filtered benchmarks containing approximately 35\% of the original instances.

\subsubsection{Selection Ratio Analysis}

\begin{figure}[t]
    \centering
    \includegraphics[width=\columnwidth]{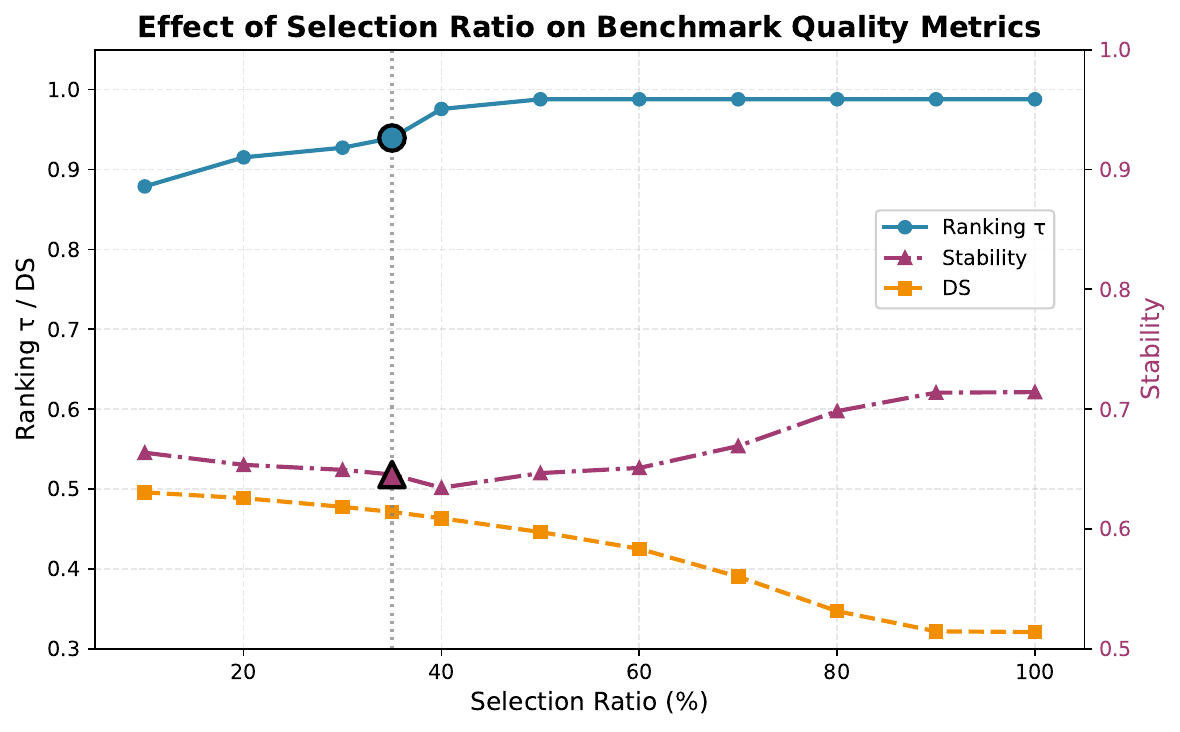}
    \caption{Effect of selection ratio on benchmark quality metrics. The optimal point at 35\% (marked) achieves a good balance between ranking consistency ($\tau$ = 0.93), stability (0.69), and discriminability (DS = 0.47).}
    \label{fig:selection_ratio}
\end{figure}

Figure~\ref{fig:selection_ratio} illustrates the effect of selection ratio on benchmark quality metrics. As the selection ratio increases from 10\% to 100\%, we observe distinct patterns across the three metrics.

Ranking consistency ($\tau$) increases rapidly from 0.88 at 10\% to 0.93 at 35\%, then plateaus near 0.99 for higher ratios. Stability shows an inverse pattern, starting high at low selection ratios (where only the most reliable instances are included) and gradually decreasing as more instances are added. DS decreases steadily as selection ratio increases, as the most discriminative instances are selected first. The optimal point at 35\% achieves a balanced trade-off: ranking $\tau$ of 0.93, stability of 0.69, and DS of 0.47—substantially better than the full benchmark's stability of 0.59 while maintaining comparable ranking consistency.

\begin{table}[t]
\centering
\small
\setlength{\tabcolsep}{3pt}
\begin{tabular}{@{}lcccc@{}}
\toprule
\textbf{Method} & \textbf{Level} & \textbf{Rank $\tau$} & \textbf{Stability} & \textbf{DS} \\
\midrule
\multicolumn{5}{c}{\textit{Single Metric}} \\
\midrule
CAD only (score > 0.15) & Inst & 0.93 & 0.61 & 0.32 \\
DS only (top 35\%) & Inst & 0.95 & 0.50 & 0.48 \\
CBRC only (top-2) & Bench & 0.62 & 0.62 & 0.28 \\
\midrule
\multicolumn{5}{c}{\textit{Two Metrics}} \\
\midrule
\textbf{CAD + DS (ours)} & \textbf{Inst} & \textbf{0.93} & \textbf{0.69} & \textbf{0.47} \\
CAD + CBRC & Mix & 0.67 & 0.78 & 0.25 \\
DS + CBRC & Mix & 0.59 & 0.58 & 0.33 \\
\midrule
\multicolumn{5}{c}{\textit{Three Metrics}} \\
\midrule
CAD + DS + CBRC & Mix & 0.65 & 0.76 & 0.29 \\
\midrule
Full benchmark & -- & 1.00 & 0.59 & 0.34 \\
\bottomrule
\end{tabular}
\caption{Metric combination ablation. Level: Inst=Instance, Bench=Benchmark. Our CAD+DS combination (bold) achieves the best balance.}
\label{tab:ablation_metrics}
\end{table}

\subsubsection{Metric Combination Ablation}

Table~\ref{tab:ablation_metrics} compares metric combinations for instance selection. Key findings from the ablation study:

\paragraph{CAD alone provides good stability.} Filtering by CAD alone (score > 0.15) yields good ranking consistency (0.93) and reasonable stability (0.61) by removing noisy instances, but reduces discriminability (0.32).

\paragraph{DS alone maximizes discriminability.} Selecting high-discriminability instances preserves the best discriminability (0.48) but provides lower stability (0.50).

\paragraph{Combined approach balances objectives.} Our CAD+DS combination achieves strong ranking consistency (0.93) and improved stability (0.69) compared to the full benchmark (0.59), while maintaining good discriminability (0.47).

\begin{table}[t]
\centering
\small
\begin{tabular}{lccc}
\toprule
\textbf{Threshold} & \textbf{Retained} & \textbf{$\tau$} & \textbf{Stab.} \\
\midrule
0.40 & 58\% & 0.87 & 1.00 \\
0.15 & 84\% & 0.93 & 0.69 \\
0.05 & 95\% & 0.96 & 0.55 \\
0.01 & 98\% & 0.98 & 0.52 \\
None & 100\% & 0.95 & 0.50 \\
\bottomrule
\end{tabular}
\caption{CAD threshold sensitivity analysis. Higher threshold is more restrictive (fewer instances retained).}
\label{tab:ablation_threshold}
\end{table}

\subsubsection{Threshold Sensitivity Analysis}

Table~\ref{tab:ablation_threshold} presents results for different CAD threshold values. The threshold analysis reveals a trade-off between ranking consistency and stability. A very restrictive threshold (0.40) achieves perfect stability (1.00) but lower ranking consistency (0.87) due to insufficient instances. The threshold of 0.15 provides an optimal balance, achieving strong ranking consistency (0.93) and good stability (0.69) while retaining 84\% of instances.

\begin{table}[t]
\centering
\small
\setlength{\tabcolsep}{3pt}
\begin{tabular}{@{}lccc@{}}
\toprule
\textbf{Strategy} & \textbf{Rank $\tau$} & \textbf{Stability} & \textbf{DS} \\
\midrule
Random selection & 0.91 $\pm$ 0.04 & 0.59 $\pm$ 0.02 & 0.34 $\pm$ 0.02 \\
High accuracy & 0.87 & 0.68 & 0.33 \\
Low accuracy & 0.79 & 0.61 & 0.26 \\
Medium difficulty & 0.93 & 0.51 & 0.47 \\
Longest instances & 0.89 & 0.59 & 0.34 \\
Shortest instances & 0.76 & 0.62 & 0.31 \\
\midrule
\textbf{CAD + DS (ours)} & \textbf{0.93} & \textbf{0.69} & \textbf{0.47} \\
Full benchmark & 1.00 & 0.59 & 0.34 \\
\bottomrule
\end{tabular}
\caption{Comparison of selection strategies. \textit{High/Low accuracy}: instances where most models succeed/fail. \textit{Medium difficulty}: moderate average accuracy. \textit{Longest/Shortest}: most/fewest tokens.}
\label{tab:ablation_baseline}
\end{table}

\subsubsection{Baseline Comparison}
Table~\ref{tab:ablation_baseline} compares our selection against baselines. High-accuracy selection provides good stability (0.68) but lower ranking consistency (0.87). Medium-difficulty achieves comparable discriminability (0.47) but lower stability (0.51). Our approach achieves the best balance of stability (0.69) and discriminability (0.47) while maintaining strong ranking consistency (0.93).

\begin{table}[t]
\centering
\small
\setlength{\tabcolsep}{3pt}
\begin{tabular}{@{}lcccc@{}}
\toprule
\textbf{Benchmark} & \textbf{DeepSeek} & \textbf{Llama} & \textbf{Qwen2.5} & \textbf{Qwen3} \\
\midrule
MATH-500 & 0.48 & 0.58 & 0.74 & 0.67 \\
AIME 2024 & 0.67 & 1.00 & 1.00 & 0.88 \\
AMC 22-24 & 0.37 & 0.58 & 0.57 & 0.43 \\
OlympiadBench & 0.54 & 0.93 & 0.91 & 0.39 \\
OmniMath & 0.61 & 0.98 & 0.84 & 0.39 \\
DROP & 0.58 & 0.76 & 0.58 & 0.62 \\
ARC & 0.80 & 0.92 & 0.87 & 0.93 \\
BBH & 0.75 & 0.63 & 0.62 & 0.62 \\
SIQA & 0.24 & 0.21 & 0.20 & 0.27 \\
CommonsenseQA & 0.57 & 0.54 & 0.59 & 0.55 \\
IFEval & 0.49 & 0.72 & 0.75 & 0.66 \\
IFBench & 0.62 & 0.52 & 0.49 & 0.44 \\
EQ-Bench & 0.60 & 0.61 & 0.54 & 0.43 \\
SuperGPQA & 0.54 & 0.39 & 0.35 & 0.44 \\
MMLU-Pro & 0.57 & 0.51 & 0.55 & 0.54 \\
\bottomrule
\end{tabular}
\caption{CAD breakdown by model family (transformed scores, higher is better). Values represent $e^{-\lambda \cdot \text{inv\_rate}}$ with $\lambda=12$, where 1.0 indicates perfect alignment.}
\label{tab:cad_breakdown}
\end{table}

\subsection{CAD Breakdown by Model Family}
Table~\ref{tab:cad_breakdown} reveals family-specific patterns. Llama achieves near-perfect CAD on several benchmarks (AIME: 1.00, OmniMath: 0.98), while Qwen3 shows higher variation across benchmarks (OlympiadBench: 0.39, EQ-Bench: 0.43). SIQA exhibits consistently low CAD across all families (0.20--0.27), indicating inherent design issues.

\section{Discussion and Future Directions}

\subsection{What Makes a High-Quality Benchmark?}

Our analysis identifies three key characteristics: (1) \textit{High discriminability}---top benchmarks (AIME, OmniMath, OlympiadBench) achieve DS > 0.7 with wide score ranges; (2) \textit{Strong capability alignment}---benchmarks with CAD > 0.6 respect within-family hierarchies and feature objective evaluation; (3) \textit{Balanced quality profile}---the highest BQS scores emerge from benchmarks balancing multiple dimensions, as exemplified by AIME 2024 (BQS = 0.79).

\subsection{Implications for Benchmark Development}

We recommend that benchmark developers: (1) target DS > 0.2 and CAD > 0.6 as minimum thresholds; (2) prefer objective evaluation criteria; (3) consider selective construction using CAD+DS metrics; and (4) monitor family-specific CAD variation as an indicator of potential biases.

\subsection{Methodological Considerations}

\paragraph{On CBRC and circularity.} Using benchmarks to evaluate benchmarks raises circularity concerns. We mitigate this by selecting widely-adopted reference benchmarks, aggregating across multiple benchmarks, and complementing CBRC with two reference-independent metrics (CAD and DS).
\paragraph{On model and CAD scope.} Our evaluation spans four model families, with held-out validation confirming generalization. CAD requires multiple model sizes within a family, limiting applicability to single-variant proprietary models, but ensures reliable capability ordering.

\section{Conclusion}

We presented \textsc{Benchmark}$^2$, a framework for evaluating LLM benchmark quality through three complementary metrics: Cross-Benchmark Ranking Consistency, Discriminability Score, and Capability Alignment Deviation. Our evaluation across 15 benchmarks and 11 models reveals significant quality variations among widely-used benchmarks, and demonstrates that selective construction can maintain evaluation fidelity using only 35\% of original instances. We hope this framework helps practitioners assess benchmark reliability. Future directions include extending to generation-based evaluations with LLM-as-judge and developing dynamic quality monitoring for benchmark degradation.

\section*{Limitations}

Our study has several limitations that suggest directions for future research. First, our evaluation focuses on three domains (mathematics, reasoning, and knowledge understanding); although our metrics are domain-agnostic by design, extending validation to additional domains such as code generation, machine translation, and dialogue systems remains important future work. Second, our analysis is restricted to text-based LLM benchmarks; as multimodal large language models become increasingly prevalent, extending our framework to vision-language, audio-language, and video understanding benchmarks represents a natural next step. Third, while our evaluation spans 11 models across four diverse families, incorporating a broader range of models including proprietary systems would enhance generalizability.

\section*{Ethics Statement}

This work involves evaluation of existing public benchmarks and models, and does not introduce new data collection or human subjects research. We use only publicly available benchmarks and evaluate models through their official APIs or publicly released weights. Our framework is intended to improve benchmark quality and thereby contribute to more reliable AI evaluation.



\bibliography{custom}

\appendix

\section{Experimental Setup Details}
\label{sec:appendix_setup}

We conduct all experiments using the \textbf{EvalScope} framework \cite{evalscope2024}, an open-source evaluation toolkit that provides standardized benchmark implementations and consistent evaluation protocols. For model deployment and inference, we utilize the \textbf{vLLM} framework \cite{kwon2023vllm}, a high-throughput serving engine optimized for large language models.

Table~\ref{tab:inference_config} summarizes the key inference parameters used across all experiments. We use greedy decoding for reproducibility and set the maximum new tokens to 16384 to accommodate long-form reasoning outputs. All experiments were conducted on NVIDIA A100 80GB GPUs, with smaller models (1.5B--8B parameters) evaluated using single-GPU deployment and larger models (32B--72B parameters) utilizing multi-GPU tensor parallelism. The complete evaluation across all 15 benchmarks and 14 models required approximately 500 GPU-hours.

\begin{table*}[t]
\centering
\small
\begin{tabular}{ll}
\toprule
\textbf{Parameter} & \textbf{Value} \\
\midrule
Temperature & 0.7 \\
Top-p & 0.8 \\
Max new tokens & 16384 \\
Tensor parallelism & Model-dependent \\
GPU memory utilization & 0.90 \\
\bottomrule
\end{tabular}
\caption{Inference configuration for all model evaluations using vLLM framework.}
\label{tab:inference_config}
\end{table*}

\section{CAD Transform Parameter Selection}
\label{sec:appendix_lambda}

The Capability Alignment Deviation (CAD) metric applies an exponential transformation to convert raw inversion rates into interpretable scores: $\text{CAD}(B_i) = e^{-\lambda \cdot \text{inv\_rate}(B_i)}$. The choice of $\lambda$ affects the sensitivity of the transformed scores to variations in raw inversion rates. We conduct a systematic analysis to select an appropriate value based on five criteria: (1) \textbf{Median Mapping}---the median raw inversion rate should map to a score in the range [0.15, 0.35]; (2) \textbf{Quality Separation}---different quality levels should exhibit meaningful score differences; (3) \textbf{Excellent Quality Reward}---benchmarks with low inversion rates (raw\_cad $< 0.03$) should receive high scores ($> 0.65$); (4) \textbf{Poor Quality Penalty}---benchmarks with high inversion rates (raw\_cad $> 0.25$) should receive low scores ($< 0.10$); and (5) \textbf{Dynamic Range}---the transformation should preserve meaningful variation across the main data distribution.

Table~\ref{tab:lambda_selection} presents the evaluation of candidate $\lambda$ values against these criteria. Based on this analysis, we select $\lambda = 12$ as it achieves the highest total score (0.68) by providing strong quality separation (0.93), perfect reward for excellent benchmarks (1.00), full penalty for poor benchmarks (1.00), and complete dynamic range preservation (1.00).

Table~\ref{tab:threshold_mapping} shows how raw inversion rates translate to CAD scores with $\lambda = 12$, providing practitioners with concrete reference points for interpreting CAD values in practice.

\begin{table*}[t]
\centering
\small
\begin{tabular}{r ccccc c}
\toprule
$\lambda$ & Median & Separation & Excellent & Poor & Range & \textbf{Total} \\
\midrule
3 & 0.00 & 1.00 & 1.00 & 0.00 & 0.90 & 0.54 \\
6 & 0.00 & 1.00 & 1.00 & 0.38 & 1.00 & 0.61 \\
9 & 0.00 & 1.00 & 1.00 & 0.70 & 1.00 & 0.66 \\
12 & 0.00 & 0.93 & 1.00 & 1.00 & 1.00 & \textbf{0.68} \\
15 & 0.00 & 0.72 & 0.70 & 1.00 & 1.00 & 0.57 \\
18 & 0.00 & 0.54 & 0.70 & 1.00 & 1.00 & 0.53 \\
24 & 0.00 & 0.30 & 0.97 & 1.00 & 1.00 & 0.52 \\
30 & 0.00 & 0.17 & 0.81 & 1.00 & 1.00 & 0.45 \\
\bottomrule
\end{tabular}
\caption{Lambda parameter selection analysis. \textbf{Median}: median raw\_cad maps to 0.15--0.35 score range; \textbf{Separation}: quality level separation; \textbf{Excellent}: excellent quality (raw\_cad $<$ 0.03) receives high score; \textbf{Poor}: poor quality (raw\_cad $>$ 0.25) receives low score; \textbf{Range}: dynamic range in main data distribution. The weighted total uses coefficients 0.30, 0.25, 0.20, 0.15, 0.10 respectively.}
\label{tab:lambda_selection}
\end{table*}

\begin{table*}[t]
\centering
\small
\begin{tabular}{cc}
\toprule
\textbf{Raw CAD} & \textbf{Transformed Score} \\
\midrule
0.00 & 1.000 \\
0.02 & 0.787 \\
0.03 & 0.698 \\
0.05 & 0.549 \\
0.08 & 0.383 \\
0.10 & 0.301 \\
0.12 & 0.237 \\
0.15 & 0.165 \\
0.20 & 0.091 \\
0.25 & 0.050 \\
0.30 & 0.027 \\
0.40 & 0.008 \\
\bottomrule
\end{tabular}
\caption{Raw CAD to transformed score mapping with $\lambda = 12$. Quality levels: Excellent (raw\_cad $<$ 0.03), Good (0.03--0.08), Acceptable (0.08--0.15), Concerning (0.15--0.25), Poor ($>$ 0.25).}
\label{tab:threshold_mapping}
\end{table*}

\section{BQS Weight and Normalization Details}
\label{sec:appendix_bqs}

The Benchmark Quality Score (BQS) combines three metrics with different native scales. To ensure meaningful aggregation, we apply normalization and empirically-tuned weights.

\paragraph{CBRC Normalization.} CBRC (Kendall's $\tau$) ranges from $-1$ to $1$, while DS and CAD both range from $0$ to $1$. To align scales, we normalize CBRC using a linear transformation:
\begin{equation}
    \widetilde{\text{CBRC}}(B_i) = \frac{\text{CBRC}(B_i) + 1}{2}
\end{equation}
This maps the CBRC range $[-1, 1]$ to $[0, 1]$, where $0$ indicates perfect negative correlation, $0.5$ indicates no correlation, and $1$ indicates perfect positive correlation.

\paragraph{Weight Selection.} We assign weights $\alpha = 0.3$, $\beta = 0.3$, and $\gamma = 0.4$ based on the following considerations:
\begin{itemize}
    \item \textbf{CAD receives the highest weight (0.4)} because it operates at the instance level and directly measures whether individual test items respect capability hierarchies—a fundamental property of well-designed benchmarks.
    \item \textbf{CBRC and DS receive equal weights (0.3 each)} as they capture complementary benchmark-level properties: external consistency (CBRC) and internal discriminative power (DS).
\end{itemize}

\paragraph{Final Formula.} The complete BQS formula is:
\begin{equation}
\begin{split}
    \text{BQS}(B_i) = &\ 0.3 \cdot \frac{\text{CBRC}(B_i) + 1}{2} + 0.3 \cdot \text{DS}(B_i) \\
    &+ 0.4 \cdot \text{CAD}(B_i)
\end{split}
\end{equation}

\section{Detailed Model Performance}
\label{sec:appendix_performance}

This section presents the complete performance matrix for all 14 models across the 15 benchmarks, organized by domain.

In the Mathematics domain (Table~\ref{tab:perf_mathematics}), DeepSeek-R1-Distill-Qwen-32B shows strong performance on competition-style benchmarks, achieving the highest score on AIME 2024 (53.3\%). The Qwen3 family demonstrates consistently strong results across all mathematics benchmarks, with Qwen3-32B achieving the highest scores on MATH-500 (87.0\%) and AMC 22-24 (67.2\%).

In General Reasoning (Table~\ref{tab:perf_general}), Qwen3-32B achieves the highest scores on DROP (85.7\%) and ARC (95.0\%). The results show clear capability hierarchies across model families, with larger models consistently outperforming their smaller counterparts.

For Knowledge \& Understanding (Table~\ref{tab:perf_knowledge}), the larger instruction-tuned models generally achieve higher performance, with IFEval and EQ-Bench showing clearer capability hierarchies across model families.

\begin{table*}[t]
\centering
\small
\begin{tabular}{lccccc}
\toprule
\textbf{Model} & \textbf{MATH-500} & \textbf{AIME 2024} & \textbf{AMC 22-24} & \textbf{OlympiadBench} & \textbf{OmniMath} \\
\midrule
DeepSeek-R1-Distill-Qwen-1.5B & 68.4 & 16.7 & 35.8 & 6.8 & 6.7 \\
DeepSeek-R1-Distill-Qwen-7B & 73.6 & 16.7 & 39.6 & 49.4 & 34.8 \\
DeepSeek-R1-Distill-Qwen-32B & 74.6 & 53.3 & 41.8 & 61.8 & 56.2 \\
Llama-3.1-Instruct-8B & 49.4 & 0.0 & 16.4 & 0.7 & 0.2 \\
Llama-3.1-Instruct-70B & 68.2 & 23.3 & 26.9 & 18.3 & 13.9 \\
Qwen2.5-Instruct-1.5B & 49.8 & 0.0 & 18.7 & 0.0 & 0.0 \\
Qwen2.5-Instruct-7B & 75.2 & 13.3 & 40.3 & 8.1 & 10.4 \\
Qwen2.5-Instruct-72B & 83.4 & 16.7 & 50.7 & 52.2 & 44.1 \\
Qwen3-1.7B & 72.2 & 10.0 & 40.3 & 28.9 & 22.9 \\
Qwen3-8B & 84.6 & 26.7 & 61.9 & 61.8 & 57.1 \\
Qwen3-32B & 87.0 & 36.7 & 67.2 & 64.8 & 62.0 \\
\bottomrule
\end{tabular}
\caption{Mathematics domain: Model performance (\%) on each benchmark.}
\label{tab:perf_mathematics}
\end{table*}

\begin{table*}[t]
\centering
\small
\begin{tabular}{lccccc}
\toprule
\textbf{Model} & \textbf{DROP} & \textbf{ARC} & \textbf{BBH} & \textbf{SIQA} & \textbf{CommonsenseQA} \\
\midrule
DeepSeek-R1-Distill-Qwen-1.5B & 42.0 & 56.7 & 30.8 & 24.9 & 37.3 \\
DeepSeek-R1-Distill-Qwen-7B & 66.2 & 83.9 & 69.4 & 37.6 & 62.4 \\
DeepSeek-R1-Distill-Qwen-32B & 76.6 & 93.2 & 89.6 & 51.4 & 83.1 \\
Llama-3.1-Instruct-8B & 66.8 & 86.3 & 67.5 & 42.3 & 76.7 \\
Llama-3.1-Instruct-70B & 87.9 & 93.9 & 85.3 & 49.6 & 82.5 \\
Qwen2.5-Instruct-1.5B & 45.9 & 78.1 & 39.7 & 41.5 & 66.5 \\
Qwen2.5-Instruct-7B & 70.2 & 90.6 & 71.1 & 52.2 & 82.4 \\
Qwen2.5-Instruct-72B & 73.7 & 94.2 & 87.3 & 52.8 & 85.4 \\
Qwen3-1.7B & 67.6 & 89.0 & 72.8 & 45.9 & 72.7 \\
Qwen3-8B & 82.0 & 94.1 & 83.5 & 48.9 & 83.0 \\
Qwen3-32B & 85.7 & 95.0 & 89.9 & 53.7 & 84.4 \\
\bottomrule
\end{tabular}
\caption{General Reasoning domain: Model performance (\%) on each benchmark.}
\label{tab:perf_general}
\end{table*}

\begin{table*}[t]
\centering
\small
\begin{tabular}{lccccc}
\toprule
\textbf{Model} & \textbf{IFEval} & \textbf{IFBench} & \textbf{EQ-Bench} & \textbf{SuperGPQA} & \textbf{MMLU-Pro} \\
\midrule
DeepSeek-R1-Distill-Qwen-1.5B & 37.1 & 11.0 & 17.1 & 9.8 & 27.1 \\
DeepSeek-R1-Distill-Qwen-7B & 58.1 & 13.1 & 49.8 & 18.0 & 43.8 \\
DeepSeek-R1-Distill-Qwen-32B & 73.6 & 19.0 & 76.8 & 31.7 & 63.0 \\
Llama-3.1-Instruct-8B & 76.7 & 25.6 & 66.7 & 20.8 & 47.6 \\
Llama-3.1-Instruct-70B & 87.2 & 32.2 & 82.1 & 35.5 & 67.4 \\
Qwen2.5-Instruct-1.5B & 41.1 & 14.5 & 48.2 & 17.3 & 27.8 \\
Qwen2.5-Instruct-7B & 74.2 & 23.6 & 72.9 & 28.9 & 57.3 \\
Qwen2.5-Instruct-72B & 86.6 & 32.7 & 78.4 & 40.5 & 71.9 \\
Qwen3-1.7B & 70.6 & 22.5 & 63.0 & 20.5 & 47.4 \\
Qwen3-8B & 84.5 & 30.6 & 76.6 & 28.0 & 61.7 \\
Qwen3-32B & 86.6 & 32.4 & 80.1 & 39.4 & 71.0 \\
\bottomrule
\end{tabular}
\caption{Knowledge \& Understanding domain: Model performance (\%) on each benchmark.}
\label{tab:perf_knowledge}
\end{table*}

\section{Statistical Reliability Analysis}
\label{sec:appendix_bootstrap}

We compute 95\% confidence intervals for all metrics using bootstrap sampling with 1000 iterations. Table~\ref{tab:bootstrap_ci} presents these intervals across all benchmarks. The results reveal that CBRC estimates show moderate uncertainty (typical CI width of 0.3--0.5), while CAD estimates are notably more stable (typical CI width $<$ 0.1). This stability arises because CAD aggregates over many instance-level comparisons, reducing variance. The DS metric shows higher variability, particularly for smaller benchmarks like AIME 2024 (CI: [0.54, 1.19]), reflecting sensitivity to the specific model set evaluated.

\begin{table*}[t]
\centering
\small
\begin{tabular}{l ccc ccc ccc}
\toprule
\multirow{2}{*}{\textbf{Benchmark}} & \multicolumn{3}{c}{\textbf{CBRC}} & \multicolumn{3}{c}{\textbf{DS}} & \multicolumn{3}{c}{\textbf{CAD}} \\
\cmidrule(lr){2-4} \cmidrule(lr){5-7} \cmidrule(lr){8-10}
 & Mean & 95\% CI & $\sigma$ & Mean & 95\% CI & $\sigma$ & Mean & 95\% CI & $\sigma$ \\
\midrule
MATH-500 & 0.76 & [0.55, 0.91] & 0.09 & 0.27 & [0.09, 0.52] & 0.12 & 0.80 & [0.76, 0.83] & 0.02 \\
AIME 2024 & 0.66 & [0.39, 0.88] & 0.13 & 0.84 & [0.54, 1.19] & 0.16 & 0.92 & [0.83, 1.00] & 0.04 \\
AMC 22-24 & 0.72 & [0.51, 0.88] & 0.09 & 0.43 & [0.24, 0.65] & 0.11 & 0.69 & [0.62, 0.75] & 0.03 \\
OlympiadBench & 0.79 & [0.61, 0.91] & 0.08 & 0.82 & [0.51, 1.15] & 0.16 & 0.77 & [0.75, 0.79] & 0.01 \\
OmniMath & 0.78 & [0.61, 0.90] & 0.07 & 0.85 & [0.58, 1.16] & 0.16 & 0.78 & [0.75, 0.80] & 0.01 \\
\midrule
DROP & 0.68 & [0.39, 0.90] & 0.13 & 0.30 & [0.11, 0.54] & 0.12 & 0.80 & [0.80, 0.81] & 0.00 \\
ARC & 0.73 & [0.51, 0.90] & 0.10 & 0.27 & [0.06, 0.51] & 0.12 & 0.94 & [0.93, 0.95] & 0.00 \\
BBH & 0.70 & [0.43, 0.90] & 0.12 & 0.31 & [0.17, 0.46] & 0.08 & 0.82 & [0.82, 0.83] & 0.00 \\
SIQA & 0.70 & [0.43, 0.90] & 0.12 & 0.29 & [0.10, 0.52] & 0.11 & 0.49 & [0.47, 0.51] & 0.01 \\
CommonsenseQA & 0.68 & [0.42, 0.88] & 0.12 & 0.28 & [0.07, 0.51] & 0.12 & 0.79 & [0.77, 0.81] & 0.01 \\
\midrule
IFEval & 0.69 & [0.47, 0.85] & 0.09 & 0.31 & [0.19, 0.45] & 0.07 & 0.79 & [0.76, 0.82] & 0.02 \\
IFBench & 0.55 & [0.22, 0.80] & 0.15 & 0.32 & [0.23, 0.39] & 0.04 & 0.74 & [0.69, 0.78] & 0.02 \\
EQ-Bench & 0.70 & [0.54, 0.85] & 0.08 & 0.35 & [0.12, 0.61] & 0.12 & 0.77 & [0.71, 0.83] & 0.03 \\
SuperGPQA & 0.65 & [0.39, 0.83] & 0.11 & 0.39 & [0.22, 0.57] & 0.09 & 0.67 & [0.67, 0.68] & 0.00 \\
MMLU-Pro & 0.59 & [0.21, 0.85] & 0.17 & 0.44 & [0.25, 0.64] & 0.10 & 0.54 & [0.53, 0.54] & 0.00 \\
\bottomrule
\end{tabular}
\caption{Bootstrap 95\% confidence intervals for all metrics across benchmarks (1000 iterations).}
\label{tab:bootstrap_ci}
\end{table*}

\section{Cross-Benchmark Correlation Analysis}
\label{sec:appendix_correlation}

We compute pairwise Kendall's $\tau$ correlations between benchmarks within each domain to understand the consistency of model rankings across different evaluation instruments.

In Mathematics (Table~\ref{tab:corr_math}), we observe high correlations among most benchmarks. MATH-500 and AMC 22-24 show strong correlation ($\tau = 0.88$), while OlympiadBench and OmniMath form a nearly perfectly correlated pair ($\tau = 0.99$). AIME 2024 shows moderate correlations with other benchmarks ($\tau \approx 0.62$--$0.71$), reflecting its unique difficulty level.

In General Reasoning (Table~\ref{tab:corr_general}), DROP and BBH show the highest correlation ($\tau = 0.85$), both requiring complex reasoning. SIQA and CommonsenseQA show strong alignment ($\tau = 0.80$), as both focus on social and commonsense understanding.

The Knowledge \& Understanding domain (Table~\ref{tab:corr_knowledge}) exhibits a relatively uniform correlation structure, with IFEval and EQ-Bench showing strong alignment ($\tau = 0.80$).

\begin{table*}[t]
\centering
\small
\begin{tabular}{lccccc}
\toprule
 & MATH-500 & AIME 2024 & AMC 22-24 & OlympiadBench & OmniMath \\
\midrule
MATH-500 & 1.00 & 0.65 & 0.88 & 0.75 & 0.75 \\
AIME 2024 & -- & 1.00 & 0.62 & 0.71 & 0.69 \\
AMC 22-24 & -- & -- & 1.00 & 0.70 & 0.70 \\
OlympiadBench & -- & -- & -- & 1.00 & 0.99 \\
OmniMath & -- & -- & -- & -- & 1.00 \\
\bottomrule
\end{tabular}
\caption{Mathematics domain: Pairwise Kendall's $\tau$ correlation between benchmarks.}
\label{tab:corr_math}
\end{table*}

\begin{table*}[t]
\centering
\small
\begin{tabular}{lccccc}
\toprule
 & DROP & ARC & BBH & SIQA & CommonsenseQA \\
\midrule
DROP & 1.00 & 0.74 & 0.85 & 0.58 & 0.56 \\
ARC & -- & 1.00 & 0.71 & 0.76 & 0.74 \\
BBH & -- & -- & 1.00 & 0.65 & 0.63 \\
SIQA & -- & -- & -- & 1.00 & 0.80 \\
CommonsenseQA & -- & -- & -- & -- & 1.00 \\
\bottomrule
\end{tabular}
\caption{General Reasoning domain: Pairwise Kendall's $\tau$ correlation between benchmarks.}
\label{tab:corr_general}
\end{table*}

\begin{table*}[t]
\centering
\small
\begin{tabular}{lccccc}
\toprule
 & IFEval & IFBench & EQ-Bench & SuperGPQA & MMLU-Pro \\
\midrule
IFEval & 1.00 & 0.75 & 0.80 & 0.62 & 0.57 \\
IFBench & -- & 1.00 & 0.54 & 0.47 & 0.43 \\
EQ-Bench & -- & -- & 1.00 & 0.80 & 0.67 \\
SuperGPQA & -- & -- & -- & 1.00 & 0.69 \\
MMLU-Pro & -- & -- & -- & -- & 1.00 \\
\bottomrule
\end{tabular}
\caption{Knowledge \& Understanding domain: Pairwise Kendall's $\tau$ correlation between benchmarks.}
\label{tab:corr_knowledge}
\end{table*}

\end{document}